\documentclass[10pt,twocolumn,letterpaper]{article}

\usepackage[pagenumbers]{cvpr}      

\newcommand{\name}{Mask-Adapter}
\usepackage{multirow} 
\usepackage{arydshln}
\usepackage{colortbl}

\definecolor{cvprblue}{rgb}{0.21,0.49,0.74}
\usepackage[pagebackref,breaklinks,colorlinks,allcolors=cvprblue]{hyperref}

\def\paperID{****} 
\def\confName{CVPR}
\def\confYear{2025}
\renewcommand{\name}{Mask-Adapter}

\begin{document}

\title{Mask-Adapter: The Devil is in the Masks for Open-Vocabulary Segmentation}

\author{
Yongkang Li$^{\star}$,
Tianheng Cheng$^{\star}$,
Bin Feng,
Wenyu Liu,
Xinggang Wang$^{\dagger}$\\
[1mm]
School of EIC, Huazhong University of Science \& Technology\\
\small\texttt{\{liyk, thch, fengbin, liuwy, xgwang\}@hust.edu.cn}}

\maketitle
\begin{abstract}
Recent open-vocabulary segmentation methods adopt mask generators to predict segmentation masks and leverage pre-trained vision-language models, \eg, CLIP, to classify these masks via mask pooling.
Although these approaches show promising results, 
it is counterintuitive that accurate masks often fail to yield accurate classification results through pooling CLIP image embeddings within the mask regions.
In this paper, we reveal the performance limitations of mask pooling and introduce \textbf{\name{}}, a simple yet effective method to address these challenges in open-vocabulary segmentation.
Compared to directly using proposal masks, our proposed \name{} extracts \textit{semantic activation maps} from proposal masks, providing richer contextual information and ensuring alignment between masks and CLIP.
Additionally, we propose a \textit{mask consistency loss} that encourages proposal masks with similar IoUs to obtain similar CLIP embeddings to enhance models' robustness to varying predicted masks.
\name{} integrates seamlessly into open-vocabulary segmentation methods based on mask pooling in a plug-and-play manner, delivering more accurate classification results. Extensive experiments across several zero-shot benchmarks demonstrate significant performance gains for the proposed \name{} on several well-established methods.
Notably, \name{} also extends effectively to SAM and achieves impressive results on several open-vocabulary segmentation datasets.
Code and models are available at \url{https://github.com/hustvl/MaskAdapter}.

\end{abstract}

\section{Introduction}
\label{sec:intro}
\begin{figure}
    \centering
    \includegraphics[width=1.0\linewidth]{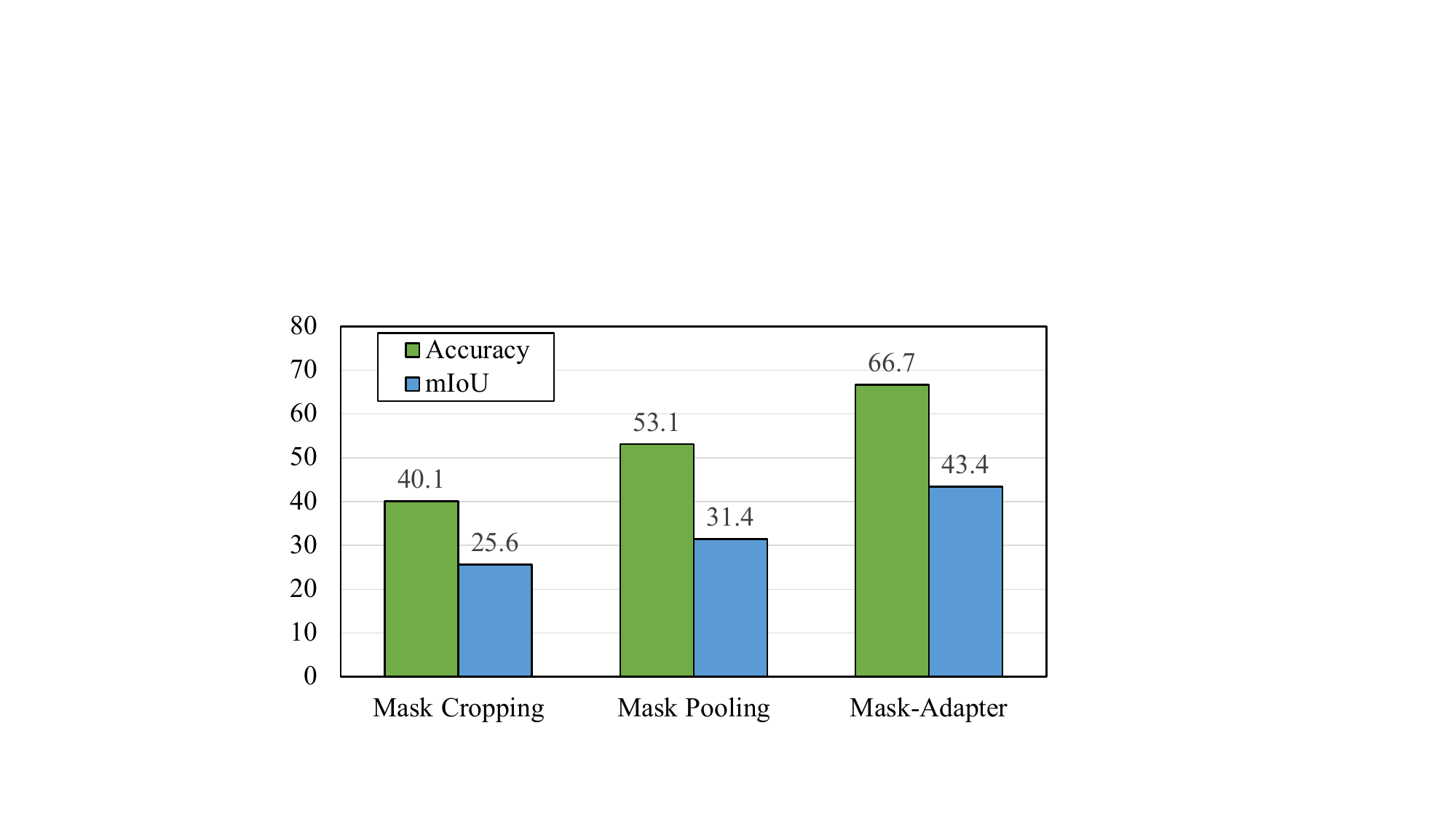}
    \vspace{-20pt}
    \caption{\textbf{Upper bound analysis of mask embedding extraction methods.} Using ADE20K ground-truth masks as input, we evaluate the upper bound of different extraction methods. Although mask cropping and mask pooling show limited performance with ground-truth masks, our Mask-Adapter significantly enhances the upper bound for open-vocabulary segmentation.}
    \label{fig:intro}
    \vspace{-10pt}
\end{figure}
\let\thefootnote\relax\footnotetext{$^\star$ Equal contribution.}
\let\thefootnote\relax\footnotetext{$^\dagger$ Corresponding author: Xinggang Wang}

Image segmentation, which aims to assign each pixel a label from a predefined set of categories, has seen significant advancements.
Despite the success of numerous image segmentation approaches~\cite{cheng2022masked,chen2018encoder,xie2021segformer,wang2020deep,zheng2021rethinking,strudel2021segmenter,cheng2022sparse}, these approaches cannot segment categories unseen during training, limiting their applicability in open-world scenarios.
Open-vocabulary segmentation (OVS) has expanded traditional image segmentation from fixed-set to open-set, enabling the segmentation of categories not appeared in the training set.
In contrast to fixed-category segmentation, open-vocabulary segmentation not only supports open-set segmentation but also enables challenging segmentation based on arbitrary text inputs, \eg, categories and descriptions.

\begin{figure*}
    \centering
    \includegraphics[width=1.0\linewidth]{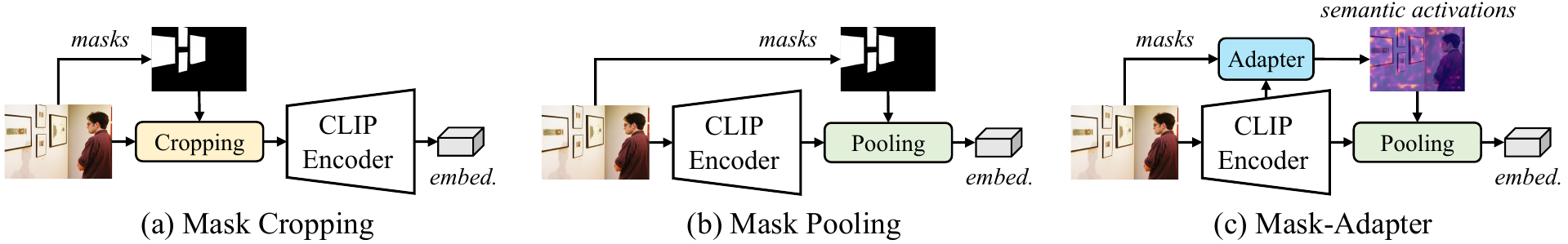}
    \vspace{-15pt}
    \caption{\textbf{Comparison of mask embedding extraction methods.} (a) \textbf{Mask Cropping}: cropping the segmented region from the image and feeding it into CLIP to extract mask embeddings. (b) \textbf{Mask Pooling}: aggregating region features with the proposal masks. (c) \textbf{Mask-Adapter}: the proposal masks and CLIP features are passed through Mask-Adapter to extract semantic activation maps, which are then used to construct mask embeddings by aggregating CLIP features based on these highlighted regions and contextual information.}
    \label{fig:com_methods}
\end{figure*}

Recent research tackles open-vocabulary segmentation by leveraging pre-trained vision-language models, \eg, CLIP~\cite{radford2021learning} and ALIGN~\cite{jia2021scaling}, which are pre-trained on large-scale image-text pairs and have strong zero-shot recognition capabilities.
Notably, several works~\cite{xu2022simple,ding2022decoupling,liang2023open,han2023open,yu2023convolutions,jiao2024collaborative,shan2024open,xu2024generalization}, such as SimpleBaseline~\cite{xu2022simple} FC-CLIP~\cite{yu2023convolutions}, follow a \textit{segment-then-recognize} paradigm for open-vocabulary segmentation, which adopts mask generators to predict proposal masks from images and then classify the proposal masks by transferring the zero-shot recognition capability from the frozen CLIP.
Recognizing segmentation masks is primarily achieved by extracting mask embeddings through the CLIP visual encoder and matching them with text embeddings.
Specifically, the mask embeddings can be obtained in two ways: (1) \textit{mask cropping}~\cite{xu2022simple,ding2022decoupling,liang2023open}: cropping the segmented region from the image and using CLIP to extract image embeddings; (2) \textit{mask pooling}~\cite{ghiasi2022scaling,yu2023convolutions,jiao2024collaborative,shan2024open,xu2024generalization}: directly aggregating the features of the region using mask pooling~\cite{ghiasi2022scaling} with the proposal masks through dot-product, which is more efficient than mask cropping and can be end-to-end optimized.
However, the upper bound of these two methods for open-vocabulary segmentation is inherently limited, as shown in Fig.~\ref{fig:intro}. The mask cropping method (Fig.~\ref{fig:com_methods} (a)) fails to consider the substantial differences between the masked image and the natural images used during CLIP~\cite{radford2021learning} pre-training, as mentioned in~\cite{liang2023open}. Meanwhile, the mask pooling method (Fig.~\ref{fig:com_methods} (b)) fails to capture semantic details and contextual information during feature aggregation, limiting its upper bound.

In this paper, we propose a simple yet effective method to address the bottlenecks in open-vocabulary segmentation. Instead of directly using predicted masks and mask pooling (or mask cropping) to extract mask embeddings, we introduce a \textit{\name{}} (Fig.~\ref{fig:com_methods} (c)) to extract \textit{semantic activation maps} from proposal masks and CLIP features, inspired by SparseInst~\cite{cheng2022sparse}. 
Mask embeddings are aggregated based on both semantic and contextual information and subsequently matched with text embeddings for mask recognition. Fig.~\ref{fig:com_methods} compares mask cropping, mask pooling, and Mask-Adapter methods. In contrast to previous approaches, Mask-Adapter offers several key advantages: (1) rather than neglecting the background, it aggregates mask embeddings across the entire image, incorporating contextual information to enrich the feature representation; (2) unlike mask pooling, which only conveys the positional information of target regions, semantic activation maps selectively highlights informative regions relevant to recognition, while suppressing less informative areas, thus enhancing feature discriminability; (3) our method preserves the generalization capabilities of CLIP during training while concurrently improving mask recognition performance. Furthermore, we introduce a mask consistency loss to enforce similar proposal masks to obtain similar CLIP mask embeddings, thereby improving robustness to variations in predicted masks. To further mitigate overfitting, we replace the Hungarian matcher with an IoU-based matcher, enabling the model to train on a broader range of proposal masks.
In addition, we propose a mixed-mask training strategy that enhances the robustness of the Mask Adapter and improves its performance on open-vocabulary segmentation tasks.

The main contributions of this paper can be summarized as follows:
\begin{itemize}
    \item We introduce the Mask-Adapter, which addresses critical bottlenecks, \ie, misalignment between pre-trained CLIP and segmentation masks, in open-vocabulary segmentation and can be integrated seamlessly into mask pooling-based methods in a plug-and-play manner, delivering more accurate classification results.
    \item We introduce a mask consistency loss and present an IoU-based matcher to reduce overfitting and improve the robustness of \name{} along with a mixed-mask training strategy.
    \item We integrate the proposed Mask-Adapter into established open-vocabulary segmentation frameworks, demonstrating substantial performance improvements and achieving state-of-the-art results on datasets such as ADE20K and Pascal-Context. Furthermore, the Mask-Adapter effectively extends to SAM without training, achieving impressive results across multiple open-vocabulary segmentation benchmarks.
\end{itemize}

\section{Related Work}
\label{sec:formatting}

\subsection{Open-Vocabulary Segmentation}
Open-vocabulary segmentation aims to segment images based on arbitrary textual descriptions rather than predefined classes, primarily categorizing into two types: per-pixel classification~\cite{li2022language,cho2024cat,xie2024sed} and mask classification~\cite{ghiasi2022scaling,xu2022simple,ding2022decoupling,han2023open,liang2023open,xu2023open,yu2023convolutions,jiao2024collaborative}. OpenSeg~\cite{ghiasi2022scaling} first employs mask pooling method to aggregate region features for mask classification.
SimpleBaseline~\cite{xu2022simple} and Zegformer~\cite{ding2022decoupling} propose two-stage frameworks which generate class-agnostic masks and feed the masked image into CLIP for classification.
Per-pixel methods, \eg, CAT-Seg~\cite{cho2024cat} and SED~\cite{xie2024sed}, adopt cost aggregation to refine the pixel-text cost volume and predict per-pixel classification outputs. PrimitiveNet~\cite{liu2024primitivenet} decomposes complex language constraints into simple primitives and employs cross-primitive and language-primitive attention mechanisms to enhance referring segmentation performance. SAN~\cite{xu2023san} attaches an adapter to CLIP, enabling mask proposal generation and recognition.
FC-CLIP~\cite{yu2023convolutions} utilizes CLIP as a shared backbone for both mask generation and mask classification. However, both mask cropping and mask pooling approaches used in previous mask classification methods to obtain mask embeddings exhibit a limited upper bound performance, as illustrated in Fig.~\ref{fig:intro}.

Several works~\cite{liang2023open, xie2024sed} point out that the bottleneck of open-vocabulary segmentation lies in CLIP-based mask classification. 
OVSeg~\cite{liang2023open} observes the significant gap between masked images and natural images when using CLIP and addresses it by adapting CLIP with masked image-category pairs. 
MaskCLIP~\cite{ding2022open} and MasQCLIP~\cite{xu2023masqclip} adopt masked attention instead of mask cropping for mask classification. MAFT~\cite{jiao2023learning} finetunes the CLIP image encoder while employing a self-distillation loss to prevent catastrophic forgetting, and MAFTP~\cite{jiao2024collaborative} further introduces content-dependent transfer to fine-tune the text representations.
Deop~\cite{han2023open} decouples mask generation and classification and employs a heatmap-based pooling method instead of mask pooling, but its open-vocabulary segmentation performance remains relatively weak. In contrast, our proposed Mask-Adapter extracts \textit{semantic activation maps} from proposal masks, providing richer contextual information and ensuring better alignment between masks and CLIP. Compared to Deop, our method differs in the following key aspects: (1) our Mask-Adapter is a standalone module that can be integrated seamlessly into open-vocabulary segmentation frameworks and SAM; (2) the structure of our Mask-Adapter differs from the heatmap decoder in Deop. Specifically, our Mask-Adapter extracts semantic activation maps from proposal masks, rather than generating heatmaps based on image features and masks, thereby capturing richer contextual and semantic information; (3) during mask-text alignment training, we introduce a mask consistency loss and employ an IoU-based matcher to enhance the model's robustness to varying predicted masks. Furthermore, we employ a two-stage training strategy, incorporating ground-truth warmup and mixed-mask training, to stabilize training and preserve generalization. Additionally, by integrating Mask-Adapter into several well-established methods, we obtain significant improvements and achieve new state-of-the-art results across multiple datasets.

\subsection{Adaptation for Vision-Language Models}
Recent advancements in vision-language pre-training models,  \eg, CLIP~\cite{radford2021learning} and ALIGN~\cite{jia2021scaling}, have demonstrated strong open-vocabulary recognition capabilities. To further enhance CLIP's performance, various adaptation strategies have been proposed. Notably, prompt tuning methods like CoOp~\citep{zhou2022learning} and CoCoOp~\citep{zhou2022conditional} replace manually crafted prompts with learnable prompts to improve domain-specific adaptation. Feature-based methods, such as CLIP-Adapter~\citep{gao2024clip} and Tip-Adapter~\citep{zhang2021tip}, introduce lightweight adapters. CLIP-Adapter fine-tunes features through a bottleneck layer, combining learned and pre-trained features, while Tip-Adapter leverages a cache model to enhance few-shot learning with minimal fine-tuning. Despite achieving good results in image classification, CLIP and these adaptations still struggle with dense prediction tasks like open-vocabulary object detection and segmentation.

To address CLIP’s limitations in dense prediction tasks, several methods have been proposed. CLIPSelf~\citep{wu2023clipself} and SILC~\citep{naeem2023silc} leverage self-distillation to align global and local features. Furthermore, Yolo-World~\citep{cheng2024yolo} integrates the CLIP text encoder with a lightweight detector for open-vocabulary object detection, while EVF-SAM~\citep{zhang2024evf} utilizes early vision-language fusion to enable text-prompted SAM for referring expression segmentation tasks. Other approaches, such as MaskCLIP~\citep{zhou2022extract}, CLIPSurgery~\citep{li2023clip}, SCLIP~\citep{wang2023sclip}, and GEM~\citep{bousselham2024grounding}, modify CLIP's attention structure to enhance its performance on dense prediction tasks. Furthermore, GBA~\citep{xu2024generalization} introduces learnable adapters to boost CLIP's generalization in open-vocabulary segmentation, while PixelCLIP~\citep{shin2024towards} improves CLIP's fine-grained pixel representation by leveraging vision foundation models (e.g., DINO~\citep{caron2021emerging}, SAM~\citep{kirillov2023segment}) and unlabeled masks. Despite these advancements, CLIP continues to underperform on open-vocabulary region recognition tasks.

\begin{figure*}
    \centering
    \includegraphics[width=1.0\linewidth]{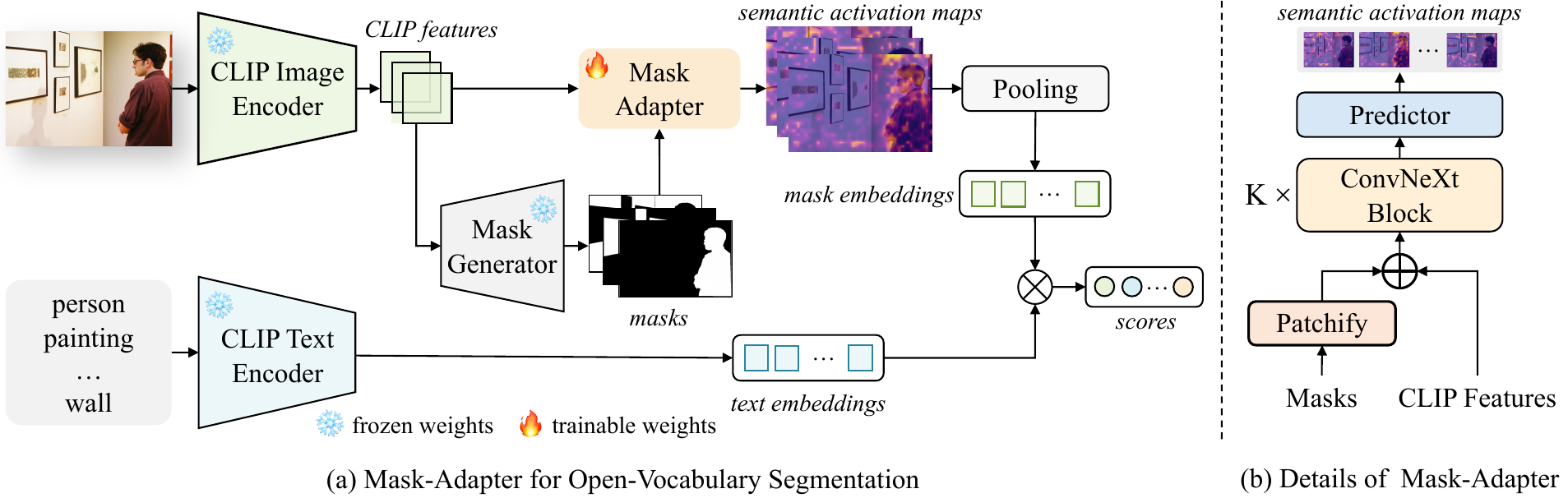}
    \vspace{-15pt}
    \caption{\textbf{Overview of \name.} (a) Mask-Adapter for Open-Vocabulary Segmentation. Mask-Adapter can be seamlessly integrated into open-vocabulary segmentation methods based on mask pooling. Mask-Adapter extracts semantic activation maps from CLIP features and proposal masks. Mask embeddings are aggregated according to semantic activation maps, which provide richer contextual and semantic information. The aggregated mask embeddings are then matched with text embeddings to perform mask classification. During training, only the Mask-Adapter is trainable. (b) Details of the Mask-Adapter. After patchifying the masks, masks and CLIP features are fused and processed through ConvNeXt blocks, ultimately outputting the semantic activation maps through a predictor.}
    \label{fig:main_arch}
    \vspace{-10pt}
\end{figure*}

\section{Approach}
\subsection{Problem Definition}
Open-vocabulary segmentation aims to segment images based on arbitrary textual descriptions, rather than a fixed set of predefined classes. This task involves training the model on a set of seen categories $\mathbf{C}_{\text{seen}}$ and evaluating its performance on both the seen categories $\mathbf{C}_{\text{seen}}$ and the unseen categories $\mathbf{C}_{\text{unseen}}$.
Existing open-vocabulary segmentation methods~\cite{xu2022simple,ding2022decoupling,liang2023open,jiao2024collaborative,yu2023convolutions} are composed of a class-agnostic mask generator and a mask classifier.
The mask generator, \eg, Mask2Former~\cite{cheng2022masked}, produces class-agnostic masks $M_p \in \mathbb{R}^{N \times H \times W} $, and the mask classifier categorizes these masks according to text descriptions.

Given \( N \) class-agnostic masks \( M_p \in \mathbb{R}^{N \times H \times W} \) generated by the mask generator, the initial approach~\cite{xu2022simple,ding2022decoupling} utilizes \textit{mask cropping} (Fig.~\ref{fig:com_methods} (a)), which crops the segmented regions from the image $\mathcal{I}$ and uses \textit{CLIP} to extract mask embeddings, as shown in the following Eq.~\ref{maskcropping}:
\begin{equation}
    \mathbf{E}_{\text{mask}} = \textit{CLIP}(\text{crop}(M_p, \mathcal{I})) \in \mathbb{R}^{N \times C}.
    \label{maskcropping}
\end{equation}
The mask embeddings are then matched with text embeddings to recognize the segmentation masks. However, the gap between masked and natural images limits the upper bound of this method for open-vocabulary segmentation.

Methods such as \cite{ghiasi2022scaling,yu2023convolutions,jiao2024collaborative} employ a \textit{mask pooling} approach (Fig.~\ref{fig:com_methods} (b)) to obtain the mask embeddings. In this method, CLIP features are aggregated via mask pooling with the proposal masks, providing greater efficiency than mask cropping, as shown in the following Eq.~\ref{maskpooling}:
\begin{equation}
    \label{maskpooling}
    \mathbf{E}_{\text{mask}} = M_p \cdot \mathcal{F}_{\text{clip}}^T.
\end{equation}
However, the mask pooling method only captures positional information of the mask region, lacking both contextual and semantic features. This limitation significantly constrains its upper bound in open-vocabulary segmentation.

\subsection{Mask-Adapter}
\paragraph{Overview.} As illustrated in Fig.~\ref{fig:main_arch}, \name{} is a simple yet remarkably effective method and can be seamlessly integrated into open-vocabulary segmentation methods to tackle the existing bottlenecks. By feeding masks and CLIP features into Mask-Adapter, it generates semantic activation maps, which highlight informative regions for each mask. With semantic activation maps, we then gather distinctive information from the CLIP features through pooling to obtain mask embeddings. Then mask embeddings are matched with text embeddings to obtain the classification results. During training, we propose ground-truth mask warmup training and mixed-mask training. At inference, masks generated by the Mask Generator are fed into Mask-Adapter for mask classification.

\paragraph{Architecture.}
Fig.~\ref{fig:main_arch} illustrates the architecture of Mask-Adapter. Given \( N \) class-agnostic masks \( M_p \), we patchify the binary masks into patches via a simple block comprising two strided \( 3 \times 3 \) convolutional layers, thus generating mask features $\mathcal{F}_{m}$ which have the same resolution as the CLIP features. We then fuse CLIP features $\mathcal{F}_{clip}$  and mask features $\mathcal{F}_{m}$ to obtain the mask region representation $\mathbf{A}_{m} \!=\! \mathcal{F}_{m} + \mathcal{F}_{clip} $. Next, $\mathbf{A}_{m}$ is passed through three ConvNeXt~\cite{liu2022convnet} blocks to enhance the semantics of the mask regions. The refined features are then processed by a final convolutional layer to generate \( K \) semantic activation maps $\mathbf{A}$ for each mask. The semantic activation maps can be expressed as follows:
\begin{equation}
    \mathbf{A} = \text{Conv}(\text{ConvNeXt}^3(\mathbf{A}_{\text{m}}))).
    \label{equ:sam_generate}
\end{equation}
We then obtain the mask embeddings \( E_{\text{m}} \) by aggregating information from the CLIP features \( \mathcal{F}_{\text{clip}} \) using \( K \) semantic activation maps \( \bar{\mathbf{A}} \) and taking the average across these embeddings, where \( \bar{\mathbf{A}} \) is normalized to 1 for each semantic map. Unlike the mask pooling method, our approach incorporates richer semantic information and includes context beyond the mask region, enhancing the upper bound for open-vocabulary segmentation. The following equation summarizes this process:
\begin{equation}
    E_{\text{m}} = \frac{1}{K} \sum_{k=1}^{K} \bar{\mathbf{A}}_k \cdot \mathcal{F}_{\text{clip}}^T, \quad E_{\text{m}} \in \mathbb{R}^{N \times C}.
    \label{equ:mask_embedding}
\end{equation}
The resulting mask embeddings serve as feature representations for the \( N \) masks, which are then matched with text embeddings to classify the segmentation masks.

\subsection{IoU-based Matcher}
Among the masks generated by the mask generator, multiple masks may correspond to the same object but be misidentified. However, the Hungarian matcher used in previous methods~\cite{stewart2016end,cheng2022masked,carion2020end}, typically finds the optimal match between ground-truth masks and predictions, which fails to capture these misclassified samples. This limitation can lead to overfitting during Mask-Adapter training. To address this, we propose an IoU-based matcher that computes the Intersection over Union (IoU) between each ground-truth mask \( y_i \) and predicted mask \( \hat{y}_j \), selecting mask pairs based on a predefined IoU threshold. This approach encourages more diverse training samples than one-to-one matching. The IoU-based matcher is defined as:
\begin{equation}
    \mathcal{M} = \{(i, j) \mid \text{IoU}(y_i, \hat{y}_j) \geq \text{IoU}_{\text{threshold}}\},
    \label{equ:iou_matching}
\end{equation}
where \( \mathcal{M} \) denotes the set of matched pairs. This method enhances the model's robustness by incorporating previously misclassified samples.

\subsection{Mask Consistency}

As discussed in~\cite{ge2025alignzeg}, allocating more space in the feature domain for unseen classes improves the recognition of unseen masks. However, directly using a standard classification loss to train the Mask-Adapter can overly crowd the feature space, limiting the model's ability to recognize unseen categories. In contrast to~\cite{ge2025alignzeg}, we adopt a simpler approach by applying a mask consistency loss to mask embeddings of seen class masks.
Given the matched predicted and ground-truth masks from the matcher, we feed them into the Mask-Adapter to obtain their respective mask embeddings \( e^{\text{gt}} \) and \( e^{\text{pred}} \). We then compute the cosine similarity loss between these embeddings as follows:
\begin{equation}
    \mathcal{L}_{\text{cos}}(e^{\text{gt}}, e^{\text{pred}}) = 1 - \sigma_{\cos}(e^{\text{gt}}, e^{\text{pred}}),
\end{equation}
where \( e^{\text{gt}} \) and \( e^{\text{pred}} \) are the embeddings of the ground-truth and predicted masks, respectively, and \( \sigma_{\cos} \) denotes cosine similarity. Mask consistency loss enforces proposal masks with similar IoUs to obtain similar CLIP embeddings, enhancing the recognition of unseen masks and improving the model's robustness.

\subsection{Stable Mask-Text Alignment Training}
To transfer CLIP's open-vocabulary recognition ability with mask-text alignment training is a natural approach for \name{}. However, directly using predicted masks leads to overfitting and training instability. To address it, we propose a two-stage training strategy including ground-truth mask warmup and mixed-mask training.

\paragraph{Ground-truth mask warmup.} We first train the Mask-Adapter using ground-truth masks to avoid low-quality predicted masks, ensuring stable training. This warmup phase enables the Mask-Adapter to develop strong generalization, making it robust to both ground-truth and predicted masks.

\paragraph{Mixed-mask training.} After the warmup training, we mix the predicted masks (from the IoU-based Matcher) and ground-truth masks for training. This mixed-mask training includes low-quality and misclassified masks, enhancing 
\name{}'s robustness and improving its performance on open-vocabulary segmentation tasks.

\paragraph{Overall training loss.} Following~\cite{cheng2022masked,yu2023convolutions}, we use cross-entropy loss for mask classification and cosine loss to enforce mask consistency. The total loss is:

\begin{equation}
    \label{loss}
    \mathcal{L} = \lambda_{ce} \cdot \mathcal{L}_{ce} + \lambda_{cos} \cdot \mathcal{L}_{cos},
\end{equation}
where $\lambda_{ce}$ and $\lambda_{cos}$ are the coefficients for the cross-entropy and cosine losses, respectively.

\begin{table*}[h!]
    \centering
    \small
    \setlength{\tabcolsep}{4pt}
    \begin{tabular}{lllc|cccccc}
        \toprule
        \textbf{Method} & \textbf{VLM} & \textbf{Training Dataset} & & \textbf{A-150} & \textbf{A-847} & \textbf{PC-59} & \textbf{PC-459} & \textbf{PAS-20} \\
        \midrule
        ZegFormer~\cite{ding2022decoupling}  & CLIP ViT-B/16 & COCO-Stuff & & 18.0 & 5.6 & 45.5 & 10.4 & 89.5  \\
        DeOP~\cite{han2023open}  & CLIP ViT-B/16 & COCO-Stuff & & 22.9 & 7.1 & 48.8 & 9.4 & 91.7  \\
        OvSeg~\cite{liang2023open}  & CLIP ViT-B/16 & COCO-Stuff & & 24.8 & 7.1 & 53.3 & 11.0 & 92.6  \\
        SAN~\cite{xu2023san}  & CLIP ViT-B/16 & COCO-Stuff & & 27.5 & 10.1 & 53.8 & 12.6 & 94.0  \\
        OpenSeg~\cite{ghiasi2022scaling}  & ALIGN & COCO Panoptic+Loc. Narr.  & & 28.6 & 8.8 & 48.2 & 12.2 & 72.2  \\
        EBSeg~\cite{shan2024open}  & CLIP ViT-B/16 & COCO-Stuff & & 30.0 & 11.7 & 56.7 & 17.3 & 94.6  \\
        SED~\cite{xie2024sed}  & CLIP ConvNeXt-B & COCO-Stuff & & 31.6 & 11.4 & 57.3 & 18.6 & 94.4  \\
        CAT-Seg~\cite{cho2024cat}  & CLIP ViT-B/16 & COCO-Stuff & & 31.8 & 12.0 & 57.5 & \textbf{19.0} & 94.6  \\
        MAFTP*~\cite{jiao2024collaborative}  & CLIP ConvNeXt-B & COCO-Stuff & & 34.5 & 13.8 & 57.5 & 18.5 & \textbf{95.5} \\
        \rowcolor{gray!10} MAFTP w/ MaskAdapter  & CLIP ConvNeXt-B & COCO-Stuff & & \textbf{35.6} & \textbf{14.2} & \textbf{58.4} & 17.9 & 95.1 \\
        \midrule
        OvSeg~\cite{ghiasi2022scaling}  & CLIP ViT-L/14 & COCO-Stuff & & 29.6 & 9.0 & 55.7 & 12.4 & 94.5  \\
        ODISE~\cite{xu2023open}  & Stable Diffusion & COCO Panoptic & & 29.9 & 11.1 & 55.3 & 14.5 & -  \\
        SAN~\cite{xu2023san}  & CLIP ViT-L/14 & COCO-Stuff & & 32.1 & 12.4 & 57.7 & 15.7 & 94.6  \\
        EBSeg~\cite{shan2024open}  & CLIP ViT-L/14 & COCO-Stuff & & 32.8 & 13.7 & 60.2 & 21.0 & 96.4  \\
        FC-CLIP*~\cite{yu2023convolutions}  & CLIP ConvNeXt-L & COCO Panoptic & & 34.1 & 14.8 & 58.4 & 18.2 & 95.4  \\
        SED~\cite{xie2024sed}  & CLIP ConvNeXt-L & COCO-Stuff & & 35.3 & 13.7 & 60.9 & 22.1 & 96.1  \\
        GBA~\cite{xu2024generalization} & CLIP ConvNeXt-L & COCO Panoptic & & 35.9 & 15.1 & 59.6 & 18.5 & 95.8  \\
        MAFTP*~\cite{jiao2024collaborative} & CLIP ConvNeXt-L & COCO-Stuff & & 36.3 & 15.5 & 59.5 & 21.2 & 96.4  \\
        \rowcolor{gray!10} FC-CLIP w/ MaskAdapter & CLIP ConvNeXt-L & COCO Panoptic & & 36.6 & 14.1 & 59.7 & 19.3 & 95.5 \\
        SMART~\cite{chng2024semantic}  & CLIP ViT-L/14 & COCO Panoptic & & 36.8 & 16.1 & 62.4 & 23.6 & -  \\
        MROVSeg~\cite{zhu2024mrovseg}   & CLIP ViT-L/14 & COCO-Stuff & & 36.9 & 16.1 & \textbf{64.1} & \textbf{24.1} & \textbf{97.6}  \\
        CAT-Seg~\cite{cho2024cat}  & CLIP ViT-L/14 & COCO-Stuff & & 37.9 & 16.0 & 63.3 & 23.8 &  97.0  \\
        \rowcolor{gray!10}MAFTP w/ MaskAdapter  & CLIP ConvNeXt-L & COCO-Stuff & & \textbf{38.2} &  \textbf{16.2} & 60.4 & 22.7 & 95.8  \\
        \bottomrule
    \end{tabular}
    \caption{\textbf{Open Vocabulary Semantic Segmentation Performance.} The best results are indicated in bold. We use mIoU as the evaluation metric. * denotes the results re-evaluated.}
\end{table*}

\section{Experiments} 
\label{expr}
In this section, we evaluate the performance of \name{} on open-vocabulary segmentation against the state-of-the-art methods and present detailed ablation studies. Additionally, we explore the potential of integrating \name{} to SAM~\cite{kirillov2023segment} for open-vocabulary segmentation.

\subsection{Experimental Setup} 
\paragraph{Baseline.} To evaluate \name{}, we mainly adopt three baseline methods, \ie, FC-CLIP (\textit{open-vocab.})~\cite{yu2023convolutions}, MAFTP (\textit{open-vocab.})~\cite{jiao2024collaborative}, and Mask2Former (\textit{fixed-vocab.})~\cite{cheng2022masked}.
FC-CLIP and MAFTP utilize mask pooling to extract mask embeddings.
In addition, we follow \cite{ghiasi2022scaling} and adapt the \textit{fixed-vocabulary} Mask2Former to classify masks with mask pooling for a fair comparison.

\vspace{-10pt} 

\paragraph{Dataset and evaluation metrics.}
Following the open-vocabulary segmentation setting~\cite{xu2022simple,liang2023open,shan2024open,cho2024cat} and baseline models~\cite{yu2023convolutions,jiao2024collaborative}, we train our Mask-Adapter separately on COCO-Stuff~\cite{caesar2018coco} and on COCO-Panoptic~\cite{lin2014microsoft}. COCO-Stuff consists of 118k images with 171 categories, while COCO-Panoptic has the same training images but contains 133 categories. Notably, training on COCO-Stuff outperforms COCO-Panoptic on open-vocabulary segmentation tasks~\cite{jiao2024collaborative}. We evaluated our model on widely-used open vocabulary segmentation benchmarks, including ADE20K~\cite{zhou2019semantic} (A-847 and A-150), Pascal-Context~\cite{mottaghi2014role} (PC-459 and PC-59), and Pascal-VOC~\cite{everingham2010pascal} (PAS-20). 
Additionally, we assess classification performance on ground-truth masks from ADE20K and explore open-vocabulary segmentation by integrating the Mask-Adapter with the Segment Anything Model~\cite{kirillov2023segment}.

We adopt mean Intersection over Union (mIoU) to evaluate semantic segmentation performance, following~\cite{xu2022simple,liang2023open,shan2024open,cho2024cat,ding2022decoupling}.
To further evaluate the model's generalization ability on unseen categories, we report mIoU\textsuperscript{s} and mIoU\textsuperscript{u} for seen and unseen categories in ablation experiments.

\vspace{-10pt} 

\begin{table}[h!]
    \centering
    \small
    \setlength{\tabcolsep}{5pt}
    \begin{tabular}{l|lll}%
        \toprule
        \multirow{2}{*}{Method w/o ensemble} & \multicolumn{3}{c}{ADE20K} \\ 
        & mIoU\textsuperscript{s} & mIoU\textsuperscript{u} & mIoU \\ 
        \midrule
        Mask2Former w/ CLIP & 34.8 & 17.5 & 26.0 \\
        Mask2Former w/ Ours & 45.3 \textcolor{red}{\scriptsize (+10.5)}  & 21.8 \textcolor{red}{\scriptsize (+4.3)} & 33.4 \textcolor{red}{\scriptsize (+7.4)} \\
        \cdashline{1-4}[1pt/1pt]
        FC-CLIP & 34.6 & 18.6 & 26.5 \\
        FC-CLIP w/ Ours & 46.2  \textcolor{red}{\scriptsize (+11.6)} & 24.8 \textcolor{red}{\scriptsize (+6.2)} & 35.4  \textcolor{red}{\scriptsize (+8.9)} \\
        \cdashline{1-4}[1pt/1pt]
        MAFTP & 45.8 & 27.0 & 36.3 \\
        MAFTP w/ Ours & 47.1  \textcolor{red}{\scriptsize (+1.3)} & 28.5  \textcolor{red}{\scriptsize (+1.5)} & 37.7  \textcolor{red}{\scriptsize (+1.4)} \\
        \bottomrule
    \end{tabular}
    \caption{\textbf{Results of baseline methods on the representative ADE20K dataset.} To more clearly highlight the effectiveness of our approach, we remove the ensemble methods.}
    \label{tab:ablation_woensemble}
    \vspace{-10pt}
\end{table}

\paragraph{Implementation details.} The proposed \name{} is built on Detectron2~\cite{wu2019detectron2} and trained in two phases. In the first phase, we train \name{} on COCO-Panoptic or COCO-Stuff using ground-truth masks for 20 epochs with a batch size of 8. 
We follow the training schedule~\cite{cheng2022masked,yu2023convolutions} and adopt the AdamW~\cite{loshchilov2017decoupled} optimizer with an initial learning rate of $1 \times 10^{-4}$, weight decay of 0.05, and a multi-step learning rate decay schedule. 
In the second phase, we employ mixed-mask training on COCO-Panoptic or COCO-Stuff for an additional 10 epochs with a batch size of 16. 
All models are trained with random flip and a crop size of 1024$\times$1024, while the evaluation is conducted with an image resolution of 896$\times$896. We set the IoU-based Matcher threshold to 0.7 and $\lambda_{\text{ce}} = 2.0$, $\lambda_{\text{cos}} = 5.0$. The \name{} generates 16 semantic activation maps for each mask. Following~\cite{yu2023convolutions}, final classification results are obtained by geometric ensembling in-vocabulary classifiers and Mask-Adapter classification results. The training was conducted on 4 NVIDIA RTX 3090 GPUs.

\begin{table}[ht]
    \centering
    \small
    \begin{tabular}{l|l|l|cc}
        \toprule
        Model & Source & Mask Embedding & Acc  \\
        \midrule
        ViT-L/14 & Open CLIP & Mask Cropping & 29.8  \\
        ConvNeXt-L & Open CLIP & Mask Cropping & 40.1  \\
        ViT-L/14 & OVSeg \cite{liang2023open} & Mask Cropping & 49.9 \\
        EVA-L/14\cite{sun2023eva} & EVA-CLIP  & Mask Pooling & 32.2 \\
        EVA-L/14\cite{sun2023eva} & CLIPSelf\cite{wu2023clipself}  & Mask Pooling & 53.1 \\
        ConvNeXt-L & Open CLIP & Mask Pooling & 53.0 \\
        ConvNeXt-L & Open CLIP & \textbf{Mask-Adapter} & 66.7  \\
        ConvNeXt-L & MAFTP & \textbf{Mask-Adapter} & \textbf{74.1}  \\
        \bottomrule
    \end{tabular}
    \caption{\textbf{Comparison with different mask embedding extraction methods for classification}. We evaluate previous approaches, including Mask Cropping and Mask Pooling methods.}
    \label{tab:ablation_mask_adapter}
    \vspace{-10pt}
\end{table}

\subsection{Main Results}
\paragraph{Open-vocabulary segmentation.} We follow standard settings~\cite{cho2024cat,xu2023san,jiao2024collaborative} to evaluate open-vocabulary segmentation. We classify the models based on the size of the Vision-Language Model (VLM). For base VLM models, Our Mask-Adapter combined with MAFTP~\cite{jiao2024collaborative} achieves the best performance on the A-150, A-847, and PC-59 datasets, with mIoU scores of 35.6, 14.2, and 58.4, respectively, improving mIoU by 1.1, 0.4, and 0.9 compared to the baseline. For large VLMs, our Mask-Adapter with MAFTP outperforms the state-of-the-art method, CAT-Seg~\cite{cho2024cat}, achieving the best results on the challenging A-150 and A-847 datasets. Compared to the baseline FC-CLIP~\cite{yu2023convolutions}, our approach improves mIoU by 2.5, 1.3, and 1.1 on the A-150, PC-59, and PC-459 datasets, respectively, demonstrating its effectiveness. Moreover, compared to the baseline MAFTP~\cite{jiao2024collaborative}, our approach improves mIoU by 1.9, 0.7, 0.9, and 1.5 on the A-150, A-847, PC-59, and PC-459 datasets, respectively, proving the adaptability of our method, which can be seamlessly integrated into open-vocabulary segmentation frameworks based on mask pooling in a plug-and-play manner.

\paragraph{Comparisons on ADE20K without ensemble strategy.}
To demonstrate the Mask-Adapter's effectiveness, we conduct experiments on ADE20K without the ensemble strategy. As shown in Tab.~\ref{tab:ablation_woensemble}, incorporating Mask-Adapter significantly boosts the performance of three baselines: Mask2Former (+28.4\%), FC-CLIP (+33.5\%), and MAFTP (+3.9\%), highlighting the substantial improvements in open-vocabulary segmentation. Additionally, we observe improvements in mIoU for both seen and unseen categories, demonstrating the strong generalization ability of our model.

\paragraph{Comparisons with different mask embedding extraction methods.}
Tab.~\ref{tab:ablation_mask_adapter} presents results of various mask embedding extraction methods on the ADE20K dataset using ground-truth masks. Our method achieves 74.1\% accuracy, outperforming the methods in \cite{xu2022simple,ding2022decoupling,liang2023open,jiao2024collaborative,yu2023convolutions} demonstrating the effectiveness of our Mask-Adapter in transferring CLIP's open-vocabulary recognition capabilities to mask classification. Additionally, rows 2, 6, and 7 directly compare mask cropping, mask pooling, and Mask-Adapter using the same model (ConvNeXt-L) and source (OpenCLIP), with Mask-Adapter outperforming the others by 66.3\% and 20.5\%, respectively.

\begin{table}[ht]
    \centering
    \small
    \setlength{\tabcolsep}{4pt}
    \begin{tabular}{cccc|ccc}
        \toprule
        \multirow{2}{*}{GT.} & \multirow{2}{*}{Pred.} &  \multirow{2}{*}{Warmup} &\multirow{2}{*}{Mask Cons.}  & \multicolumn{3}{c}{ADE20K} \\ 
        & & & & mIoU\textsuperscript{s} & mIoU\textsuperscript{u} & mIoU \\
        \midrule
         &  &  &  & 34.6 & 18.6 & 26.5 \\
         \checkmark&  &  & & 44.7 & 23.9 & 34.1 \\
         \checkmark& \checkmark &  & & 45.4 & 23.3 & 34.3 \\
         \checkmark& \checkmark & \checkmark & & 45.6 & 23.8 & 34.6 \\
         \checkmark& \checkmark & \checkmark &\checkmark &\textbf{46.2} & \textbf{24.8} & \textbf{35.4} \\
        \bottomrule
    \end{tabular}
    \caption{\textbf{Ablation study on different training strategies and methods in Mask-Adapter.} GT. refers to ADE20K ground truth masks, Pred. denotes FC-CLIP predicted masks, Mask Cons. indicates whether mask consistency is applied, and GT Warmup refers to the warmup using ground truth masks. We remove the ensemble operation here.}
    \label{tab:ablation_mask_adapter_training}
    \vspace{-10pt}
\end{table}

\subsection{Ablation Experiments}

\paragraph{Ablation study on Mask-Adapter.}
As shown in Tab.~\ref{tab:ablation_mask_adapter_training}, replacing Mask Pooling with the ground-truth trained Mask-Adapter in FC-CLIP improves mIoU by 10.1 and 5.3 for seen and unseen categories, respectively. However, directly using mixed predicted and ground-truth masks for training slightly reduces performance on unseen classes. Introducing Ground-truth warmup and mask consistency loss improves mIoU by 0.3 and 0.8, respectively, highlighting the importance of Ground-Truth Warmup and Mask Consistency in enhancing the model's robustness.


\begin{table}[ht]
    \centering
    \small
    \setlength{\tabcolsep}{5pt}
    \begin{tabular}{l|c|ccc}
        \toprule
         &  & \multicolumn{3}{c}{ADE20K} \\
        Matcher& Threshold & mIoU\textsuperscript{s} & mIoU\textsuperscript{u} & mIoU \\
        \midrule
        Hungarian  & - & 47.4 & 24.6 & 35.9 \\
        IoU-based  & 0.7 & \textbf{47.8} & \textbf{25.0} & \textbf{36.2}  \\
        IoU-based  & 0.8 & 47.2 & 24.7 & 35.8 \\
        \bottomrule
    \end{tabular}
    \caption{\textbf{Ablation study of different matchers and IoU thresholds.} We evaluate the impact of various matchers and IoU thresholds on the ADE20K dataset. The experiments are conducted during the mixed-mask training phase, without using mask consistency loss.}
    \label{tab:ablation_matcher}
    \vspace{-10pt}
\end{table}

\begin{figure*}[htbp]
    \centering
    \includegraphics[width=1\linewidth]{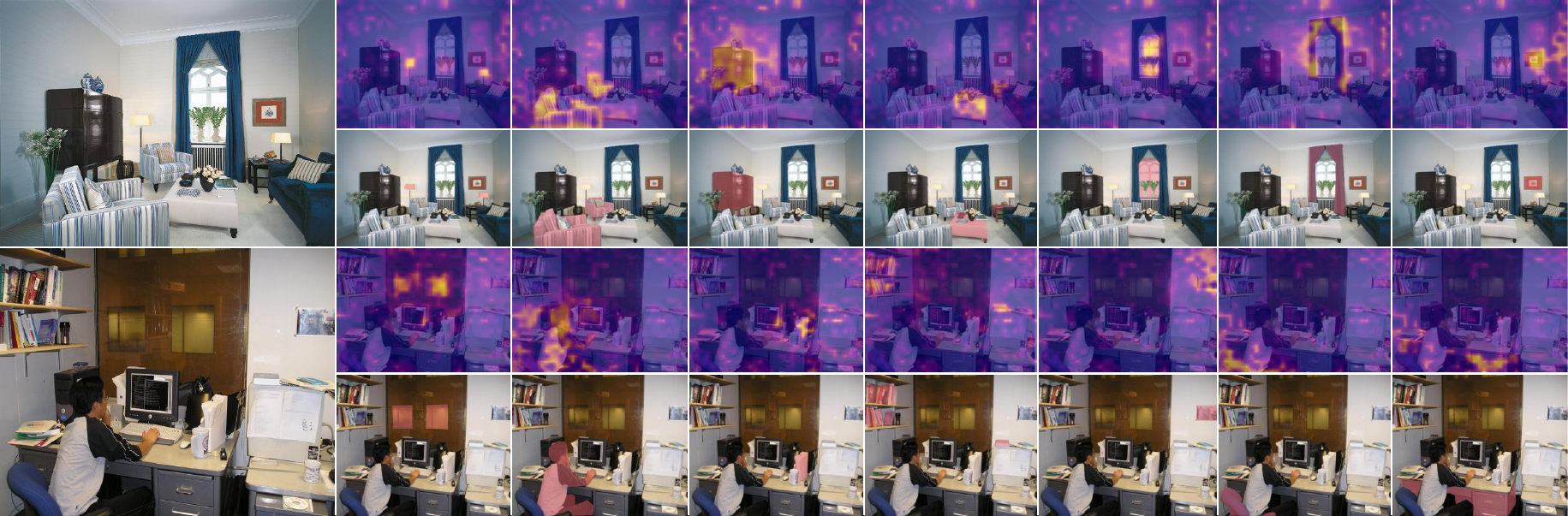}
    \vspace{-5pt}
    \caption{\textbf{Visualizations of Semantic Activation Maps.} We present visualizations of the semantic activation maps and their corresponding segmentation masks. For each input image, the top row shows the semantic activation maps, while the bottom row displays the segmentation masks. The semantic activation maps emphasize the most discriminative regions of the mask. Best viewed on screen after zooming in.}
    \label{fig:vis_sam_mask}
    \vspace{-5pt}
\end{figure*}

\paragraph{Ablation study of different matchers and IoU thresholds.} 
As shown in Tab.~\ref{tab:ablation_matcher}, we evaluate the impact of different matchers on the ADE20K dataset. The IoU-based matcher outperforms the Hungarian matcher in terms of mIoU for unseen classes. With an IoU-based matcher threshold of 0.7, the model improves by 0.3 mIoU overall, and 0.4 mIoU for seen and unseen classes.

\begin{table}[ht]
    \centering
    \small
    \setlength{\tabcolsep}{4pt}
    \scalebox{1.0}{
    \begin{tabular}{l|c|ccc}
        \toprule
         &\centering Patch Embedding & GT mIoU & FC-CLIP mIoU  \\
        \midrule
        \multirow{3}{*}{\centering (a)} 
        & Concat & 42.5 & 35.2 \\
        & Add + $2 \times 3 \times 3$ Conv  & \textbf{43.4} & \textbf{35.4} \\
        & Add + $4 \times 4$ Conv & \textbf{43.4} & 35.2 \\
        \midrule
        & \centering Num Blocks  & GT mIoU & FC-CLIP mIoU \\
        \midrule
        \multirow{3}{*}{\centering (b)} 
        & 2 &  43.1 & \textbf{35.4}  \\
        & 3 & 43.4 & \textbf{35.4}  \\
        & 4 & \textbf{43.7} & 35.3  \\
        \midrule
        & \centering CLIP Feature  & GT mIoU & FC-CLIP mIoU \\
        \midrule
        \multirow{2}{*}{\centering (c)} 
        & with  & \textbf{43.4} & \textbf{35.4}\\
        & w/o  & 35.5 & 32.5  \\
        \bottomrule
    \end{tabular}}
    \caption{\textbf{Ablation study on different designs in Mask-Adapter}, analyzing the impact of different patch embedding methods, ConvNeXt block counts, and input CLIP feature usage. Section (a) compares patch embedding strategies, (b) examines the effect of varying ConvNeXt block numbers, and (c) highlights the importance of including CLIP features. The experiments are conducted during the warm-up training phase.}
    \label{tab:patch_embedding_cnext_clip_feature}
\end{table}

\vspace{-6pt} 

\paragraph{Ablation study on different designs in Mask-Adapter.}
Tab.~\ref{tab:patch_embedding_cnext_clip_feature} presents an ablation study on key Mask-Adapter design choices. In (a), we show that adding CLIP features to mask features outperforms simple concatenation, proving this fusion strategy is the most effective. Our mask patchify approach, with two 3$\times$3 convolutions, outperforms a single 4$\times$4 convolution. In Tab.~\ref{tab:patch_embedding_cnext_clip_feature} (b), we find that increasing ConvNeXt block count improves mask classification accuracy with ground-truth masks, but too many blocks cause overfitting. Three blocks strike the optimal balance. In Tab.~\ref{tab:patch_embedding_cnext_clip_feature} (c), incorporating CLIP features significantly improves both Mask Accuracy and mIoU, enhancing semantic representation.

\paragraph{Ablation of ground-truth masks warmup datasets.}
Tab.~\ref{tab:ablation_gt_training_data} evaluates the impact of different dataset combinations for ground-truth warmup training in open-vocabulary segmentation. While COCO-Panoptic is a standard choice, we observed overfitting with this dataset, leading to reduced accuracy for unseen categories over time. To address this, we incorporate additional datasets. LVIS enhances Mask-Adapter’s robustness, likely due to its richer textual information. However, integrating a subset of the Grand dataset~\cite{rasheed2024glamm} (2\% of the SA-1B \cite{kirillov2023segment}) led to performance degradation compared to the COCO-only baseline, likely due to mask quality problems.

\begin{table}[h]
    \centering
    \small
    \setlength{\tabcolsep}{5pt}
    \begin{tabular}{l|ccc}
        \toprule
        Datasets & GT mIoU & FC-CLIP mIoU \\
        \midrule
        COCO-P & 43.4 & 35.4\\
        COCO-P, LVIS & \textbf{44.5} & \textbf{36.4} \\
        COCO-P, LVIS, Grand & 43.6 & 35.1 \\
        \bottomrule
    \end{tabular}
    \caption{\textbf{Ablations on ground-truth warmup training data.} Comparison of open-vocabulary segmentation performance (mIoU) on ADE20K, utilizing Ground Truth masks and FC-CLIP predicted masks under different dataset combinations.}
    \label{tab:ablation_gt_training_data}
\end{table}

\begin{table}[ht]
    \centering
    \small
    \setlength{\tabcolsep}{5pt}
    \begin{tabular}{l|cccc}
        \toprule
         & A-847 & A-150 & PC-459 & PC-59 \\
        \midrule
        SAM w/ CLIP & 7.1 & 17.9 & 6.4 & 34.4 \\
        SAM w/ MAFT & 10.1 & 29.1 & 12.8 & \textbf{53.5} \\
        SAM w/ Mask-Adapter & \textbf{11.4} & \textbf{31.5} & \textbf{16.8} & 49.5 \\
        \bottomrule
    \end{tabular}
    \caption{\textbf{Comparison of SAM and SAM with Mask-Adapter under open-vocabulary settings.} Following~\cite{jiao2023learning}, we evaluate performance using SAM-H with a standard CLIP model and MAFT~\cite{jiao2023learning}, as well as our Mask-Adapter approach.} 
    \label{tab:comparison_sam}
\end{table}

\subsection{Segment Anything with Mask-Adapter}
Mask-Adapter can be seamlessly integrated into SAM without training, significantly enhancing its performance on open-vocabulary segmentation tasks. Following \cite{jiao2023learning}, we conduct experiments under \textit{open-vocabulary settings}. Tab. \ref{tab:comparison_sam} shows that SAM with the Mask-Adapter outperforms SAM w/ CLIP or w/ MAFT across multiple datasets, highlighting the Mask-Adapter's effectiveness in improving the classification accuracy of masks generated by SAM.

\section{Conclusion}
In this work, we reveal severe limitations of prevalent methods using mask pooling and predicted masks for open-vocabulary segmentation and present a simple yet effective \name{}.
Instead of directly using proposal masks, \name{} extracts semantic activation maps from masks and obtains mask embeddings by aggregating CLIP features according to the semantic and contextual information provided by the semantic activation maps for mask classification. We integrate \name{} into several well-established open-vocabulary segmentation methods, demonstrating significant performance gains and achieving state-of-the-art results across multiple datasets. \name{} also extends to SAM without training and achieves impressive results on several open-vocabulary segmentation benchmarks.

\maketitlesupplementary

\renewcommand{\thesection}{\Alph{section}}
\setcounter{section}{0} 

\section{Additional Details}

\subsection{Geometric Ensemble Strategy}

Following~\cite{yu2023convolutions,ghiasi2022scaling,xu2023open}, we employ a geometric ensemble strategy to fuse the class probabilities predicted by Mask2Former, denoted as \( \hat{y}_{\text{in}} \)~\cite{yu2023convolutions}, and those predicted by Mask-Adapter, denoted as \( \hat{y}_{\text{out}} \). The geometric ensemble is defined as:

\[
y(c) = 
\begin{cases}
    \left( \hat{y}_{\text{in}}(c) \right)^{1 - \alpha} \cdot \hat{y}_{\text{out}}(c)^{\alpha}, & \text{if } c \in C_{\text{seen}} \\
    \left( \hat{y}_{\text{in}}(c) \right)^{1 - \beta} \cdot \hat{y}_{\text{out}}(c)^{\beta}, & \text{if } c \in C_{\text{unseen}}
\end{cases}
\]

Here, \( \alpha \) and \( \beta \) are hyperparameters that control the relative contributions of predictions for seen (\( C_{\text{seen}} \)) and unseen (\( C_{\text{unseen}} \)) categories. For FC-CLIP, we set \( \alpha = 0.7 \) and \( \beta = 0.9 \). For MAFTP-Base, \( \alpha = 0.7 \) and \( \beta = 1.0 \), while for MAFTP-Large, \( \alpha = 0.8 \) and \( \beta = 1.0 \). This geometric ensemble effectively balances the strengths of both predictions, enhancing the model's recognition capability.

\subsection{Datasets}

\noindent\textbf{COCO-Stuff}~\cite{caesar2018coco} extends COCO with fine-grained annotations for 80 thing and 91 stuff classes, covering 118k training images. It is commonly used for training open-vocabulary segmentation models.

\noindent\textbf{COCO-Panoptic}~\cite{lin2014microsoft} is derived from COCO and designed for panoptic segmentation, combining instance and semantic segmentation. It includes 118k training images with 133 categories: 80 thing and 53 stuff classes, a subset of COCO-Stuff's 91 stuff classes. Due to the smaller number of stuff classes, training on COCO-Panoptic yields lower performance compared to COCO-Stuff for open-vocabulary segmentation.

\noindent\textbf{ADE20K-150 (A-150)}~\cite{zhou2019semantic} is a large-scale scene parsing dataset with 20,210 training and 2,000 validation images, annotated with 150 classes. A-150 serves as the primary evaluation dataset for open-vocabulary segmentation and ablation studies. We manually classify the categories in A-150 into seen and unseen groups. Specifically, if a supercategory in ADE20K has a corresponding category in COCO, such as the ADE20K supercategory field corresponding to the COCO category playing field, we classify it as seen. This approach enables precise evaluation of the segmentation performance on unseen categories.

\noindent\textbf{ADE20K-847 (A-847)}~\cite{zhou2019semantic} contains the same 20,210 training and 2,000 validation images as A-150, with annotations for 847 categories.

\noindent\textbf{PASCAL VOC (PAS-20)}~\cite{everingham2010pascal} includes 1,464 training and 1,449 test images annotated with 20 object classes.

\noindent\textbf{PASCAL-Context (PC-59)}~\cite{mottaghi2014role} extends PASCAL VOC with 4,998 images annotated for 59 categories.

\noindent\textbf{PASCAL-Context (PC-459)}~\cite{mottaghi2014role} further extends PC-59 with annotations for 459 categories using the same 4,998 images.

\section{Additional Results}

\paragraph{Ablation of mask consistency.} 
Tab.~\ref{tab:ablation_cosine_loss} presents the effect of varying cosine loss weights in combination with the IoU-based matcher (threshold 0.7). With a cosine loss weight of 5, we observe a 0.4 improvement in overall mIoU and a 0.5 increase in mIoU for unseen classes. 

\begin{table}[ht]
    \centering
    \small
    \begin{tabular}{c|cccc}
        \toprule
        \multirow{2}{*}{Cosine Loss Weight} & \multicolumn{3}{c}{ADE20K} \\ 
        & mIoU\textsuperscript{s} & mIoU\textsuperscript{u} & mIoU \\
        \midrule
        0.0 & 47.8 & 25.0 & 36.2  \\
        2.0 & 47.8 & 24.7 & 36.1  \\
        5.0 & \textbf{48.0} & \textbf{25.5} & \textbf{36.6} \\
        10.0 & 47.4 & 24.8 & 36.0  \\
        \bottomrule
    \end{tabular}
    \caption{\textbf{Ablation study of cosine loss weight in mask consistency.} We evaluate the effect of different cosine loss weights on the ADE20K dataset combining FC-CLIP with Mask-Adapter.}
    \label{tab:ablation_cosine_loss}
    \vspace{-10pt}
\end{table}

\begin{figure}[ht]
    \centering
    \includegraphics[width=1.0\linewidth]{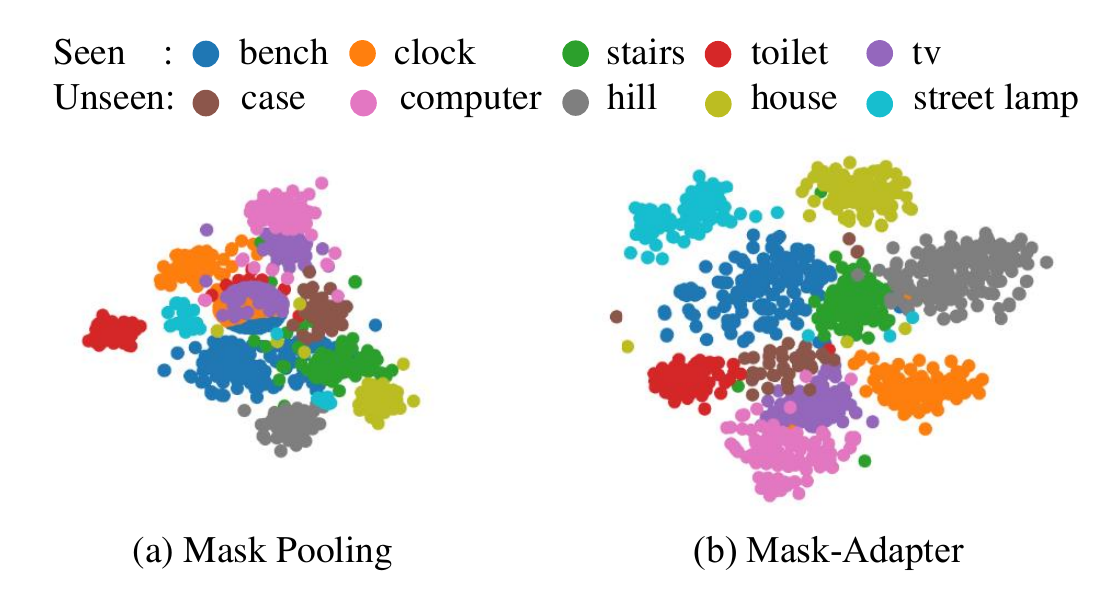} 
    \vspace{-10pt}
    \caption{\textbf{\textit{t}-SNE visualization of mask embeddings from different extraction methods.} Mask embeddings extracted using the Mask-Adapter demonstrate better separability compared to those obtained by mask pooling.}
    \label{fig:tsne_result}
    \vspace{-10pt}
\end{figure}

Fig.~\ref{fig:tsne_result} illustrates the effects of different mask extraction methods. The adoption of mask consistency constraint increases inter-class distances and enhances the distinctiveness of mask embeddings, thereby improving the model's ability to recognize unseen classes.

\begin{figure*}[ht]
    \centering
    \includegraphics[width=1.0\linewidth]{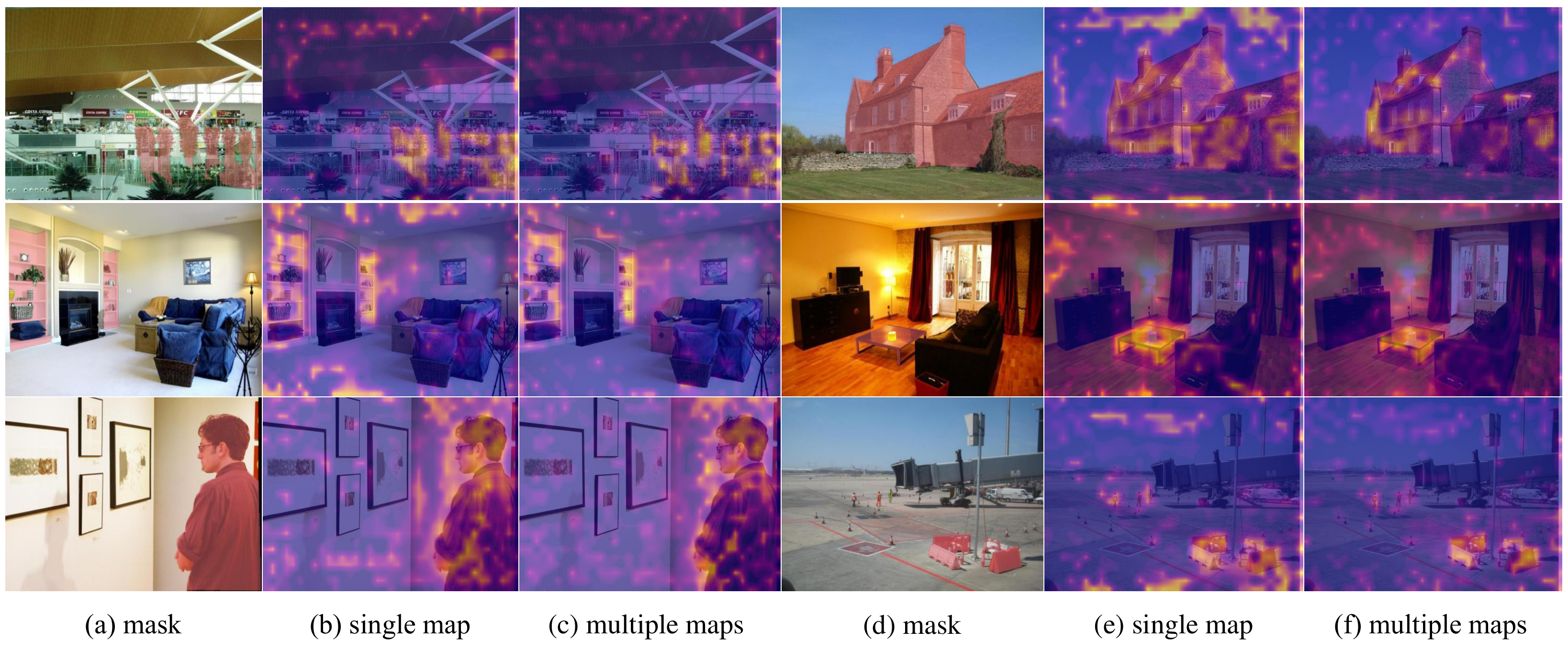} 
    \caption{\textbf{Visualization of different numbers of semantic activation maps.} We compare the visualization of a single semantic activation map and multiple semantic activation maps. Using multiple semantic activation maps effectively reduces excessive contextual noise, enhancing the model's recognition capability for masks.}
    \label{fig:maps_num}
    \vspace{-10pt}
\end{figure*}

\paragraph{Warmup training with additional datasets.} 

\begin{figure}[ht]
    \centering
    \includegraphics[width=1.0\linewidth]{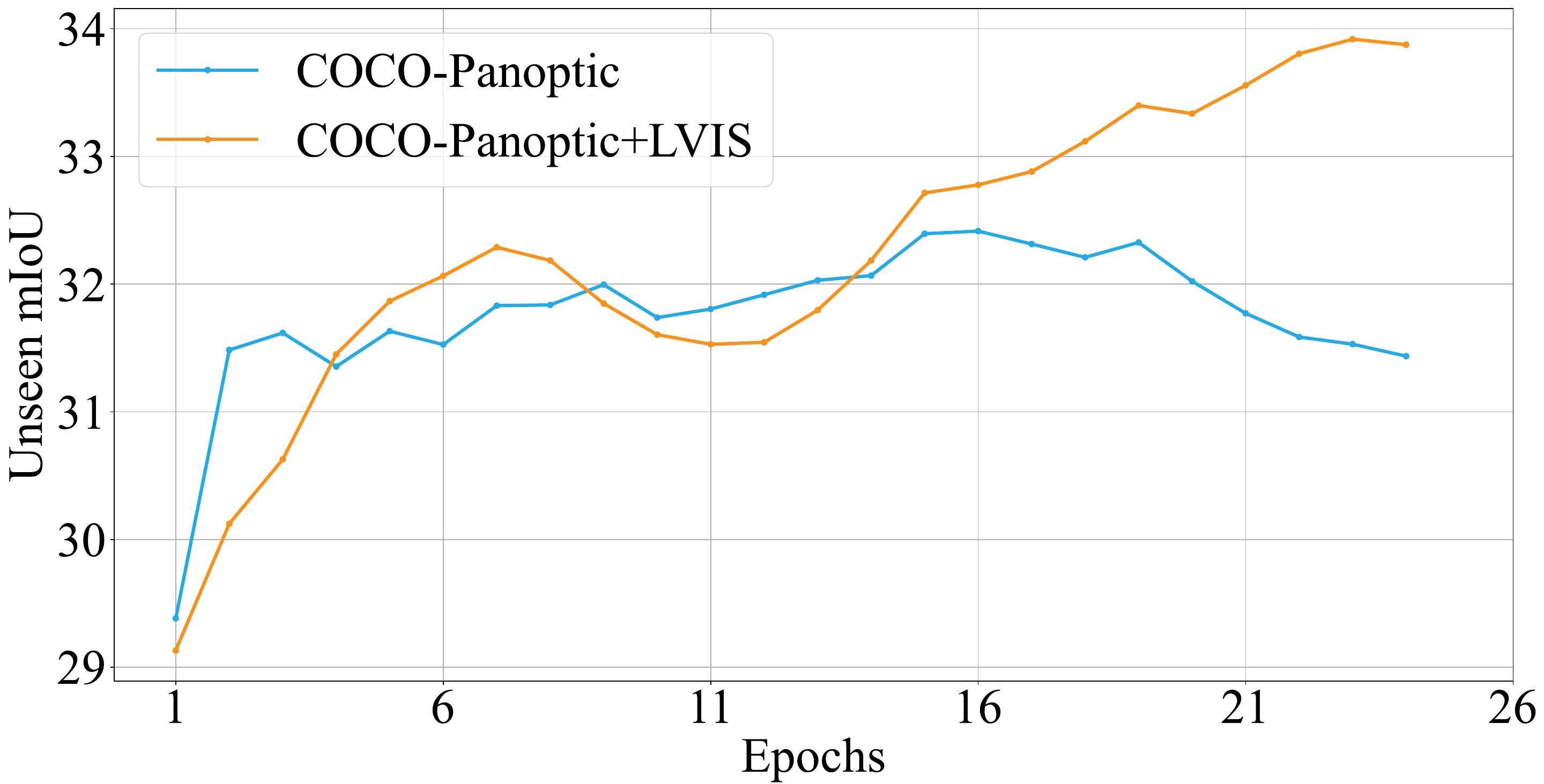} 
    \vspace{-10pt}
    \caption{\textbf{Ground-truth warmup training with different datasets.} We compare the results of training on COCO-Panoptic alone and COCO-Panoptic with additional LVIS data. The unseen mIoU is evaluated every 10,000 iterations.}
    \label{fig:dataset_com}
    \vspace{-10pt}
\end{figure}

Fig.~\ref{fig:dataset_com} demonstrates the model's performance for unseen categories using different datasets during ground-truth warmup training. Training on COCO-Panoptic often results in overfitting, impairing the model's ability to recognize unseen classes. In contrast, training on a combined COCO-Panoptic and LVIS dataset enhances the model's ability to recognize unseen categories and improves generalization, while maintaining the same number of training epochs. This improvement is primarily due to the richer categories in LVIS compared to COCO-Panoptic, which enhances the model's open-vocabulary recognition capability and reduces overfitting. This observation highlights a limitation in the current open-vocabulary segmentation setup, where training solely on COCO-Panoptic or COCO-Stuff restricts the model's generalization ability for unseen categories.

\paragraph{Ablations on the number of semantic activation maps.}

In our Mask-Adapter, we extract 16 semantic activation maps for each mask, aggregate their corresponding CLIP features separately, and compute the average. This design effectively mitigates the excessive contextual noise in the semantic activation maps, as shown in Fig.~\ref{fig:maps_num}. We also evaluate the impact of using different numbers of semantic activation maps during ground-truth warmup training in Tab.~\ref{tab:ablation_num_maps}. Compared to a single semantic activation map, multiple semantic activation maps reduce the excessive contextual noise and improve the FC-CLIP mIoU by 0.8, demonstrating the effectiveness of our method in enhancing the model's recognition capability and robustness.

\begin{table}[h]
    \centering
    \small
    \setlength{\tabcolsep}{5pt}
    \begin{tabular}{c|ccc}
        \toprule
        num. of maps & GT mIoU & FC-CLIP mIoU\\
        \midrule
        1 & 41.9 & 34.6   \\
        16 & \textbf{43.4} & \textbf{35.4}   \\
        \bottomrule
    \end{tabular}
    \caption{\textbf{Comparison with different numbers of semantic activation maps during ground-truth warmup.} GT mIoU and FC-CLIP mIoU represent results using Ground-truth and FC-CLIP predicted masks, respectively.}
    \label{tab:ablation_num_maps}
\end{table}

\paragraph{Ablations on the block structures.}
\begin{table}[h]
    \small
    \centering
    \setlength{\tabcolsep}{5pt}
    \begin{tabular}{l|cc}
            \toprule
            Block & GT mIoU & FC-CLIP mIoU \\
            \midrule
            ResNet Block & 43.0 & 35.1  \\
            ConvNeXt Block& \textbf{43.4} & \textbf{35.4}   \\
            Transformer Block & \textbf{43.4} & 35.0   \\
            Swin Transformer Block & 42.6 & 34.9   \\
            \bottomrule
        \end{tabular}
        
        \caption{\textbf{Ablation on different blocks structures.}}
    \label{tab:diff_blocks}
\end{table}

In our paper, we primarily adopt ConvNeXt blocks, which are consistent with the backbone. We also provide the ablations about different blocks in Tab.~\ref{tab:diff_blocks}. The results demonstrate that the ConvNeXt structure yields superior performance compared to other block configurations, and CNN-based blocks are generally more suitable for dense prediction tasks.


\begin{figure*}[ht]
    \centering
    \includegraphics[width=1.0\linewidth]{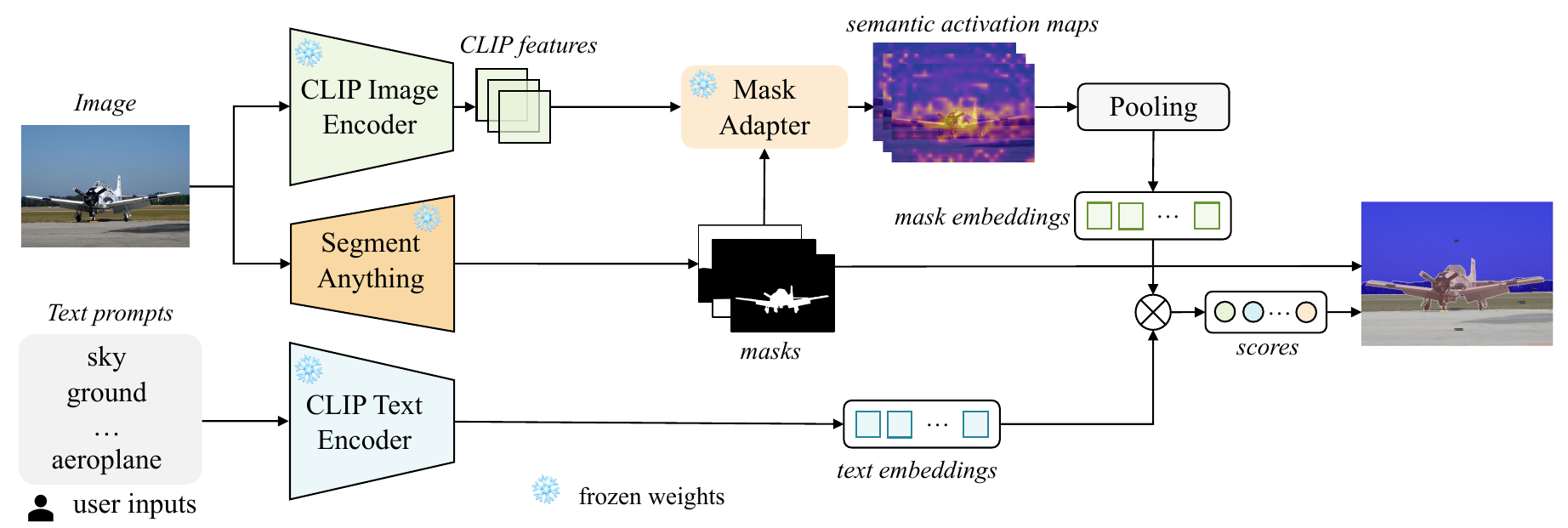}
    \vspace{-15pt}
    \caption{\textbf{Illustration of the framework for Segment Anything with Mask-Adapter.} The SAM generates class-agnostic masks, while CLIP extracts features. These are processed by the Mask-Adapter to produce semantic activation maps and obtain mask embeddings, which are matched with text embeddings for classification.}
    \label{fig:sam_withadapter}
    \vspace{-10pt}
\end{figure*}

\paragraph{Experimental analysis on PASCAL VOC.}

As mentioned in~\cite{xu2023san}, PASCAL VOC and PASCAL-Context(PC-59) exhibit a Hausdorff similarity of approximately 0.9 with COCO-Stuff. similarity, all categories in PASCAL VOC overlap with those in COCO-Stuff, which limits its ability to evaluate the model's performance on unseen classes. Mask-Adapter is trained to extract semantic activation maps, aiming to retain the original generalization capabilities of CLIP while improving mask classification accuracy. To demonstrate the effectiveness of our Mask-Adapter over original Mask-Pooling, we evaluate PASCAL VOC (PAS-20) using the predicted probabilities from Mask2Former, mask pooling, and Mask-Adapter independently.
\begin{table}[ht]
\centering
   \small
   \setlength{\tabcolsep}{10pt}
    \begin{tabular}{l|c}
    \toprule
        Classification      &  PASCAL VOC mIoU \\ \hline
    \midrule
        In-vocab.    & 94.6                   \\ 
        Out-vocab.   & 92.1                   \\
        In-vocab. \& Out-vocab.     & 95.4                   \\ 
        Mask-Adapter    & 95.4                   \\ 
        In-vocab. \& Mask-Adapter    & 95.5                   \\ 
        \bottomrule
    \end{tabular}
\caption{\textbf{Comparison of class branches on PASCAL VOC.} In-vocab. represents the results obtained using the predictions from Mask2Former, while Out-vocab. refers to the results using the predictions from Mask-Pooling.}

\label{tab:class_branch_pascal_voc}
\end{table}
From the results in Tab.~\ref{tab:class_branch_pascal_voc}, we observe that Mask-Adapter improves upon the In-vocab. and Out-vocab. branches by 0.8 and 3.3, respectively. This indicates that our model demonstrates substantial improvements for both seen and unseen categories. However, due to the limited number of categories in Pascal VOC, \ie, 20 common categories, the overall improvement brought by Mask-Adapter is somewhat minor. 

\section{Extending to Segment Anything}

Our Mask-Adapter can be seamlessly integrated into SAM~\cite{kirillov2023segment} in a plug-and-play manner, recognizing class-agnostic masks predicted by SAM. As illustrated in Fig.~\ref{fig:sam_withadapter}, given an input image, SAM generates class-agnostic masks, while CLIP extracts image features. These outputs are fed into the Mask-Adapter to extract semantic activation maps. Subsequently, the CLIP features are aggregated to generate mask embeddings, which are matched with text embeddings for mask classification.

In our experiments, we utilize SAM-H for mask generation and CLIP ConvNeXt-L~\cite{liu2022convnet} for feature extraction. We use SAM's default \texttt{AutomaticMaskGenerator}, which first samples points from the image and then uses these points as prompts to perform segmentation on the image. Without any additional training or fine-tuning, our Mask-Adapter combined with SAM achieved a remarkable mIoU of 31.4 on the A-150 dataset, surpassing the performance of ODISE~\cite{xu2023open}. This demonstrates the adaptability and effectiveness of our approach. However, one challenge of SAM lies in its exceptionally fine-grained mask outputs, which can negatively impact its performance on open-vocabulary semantic segmentation tasks. Addressing this limitation remains an open problem and is beyond the scope of this paper.

\section{Visualizations}
\begin{figure*}[htbp]
\centering
\includegraphics[width=1.00\linewidth]{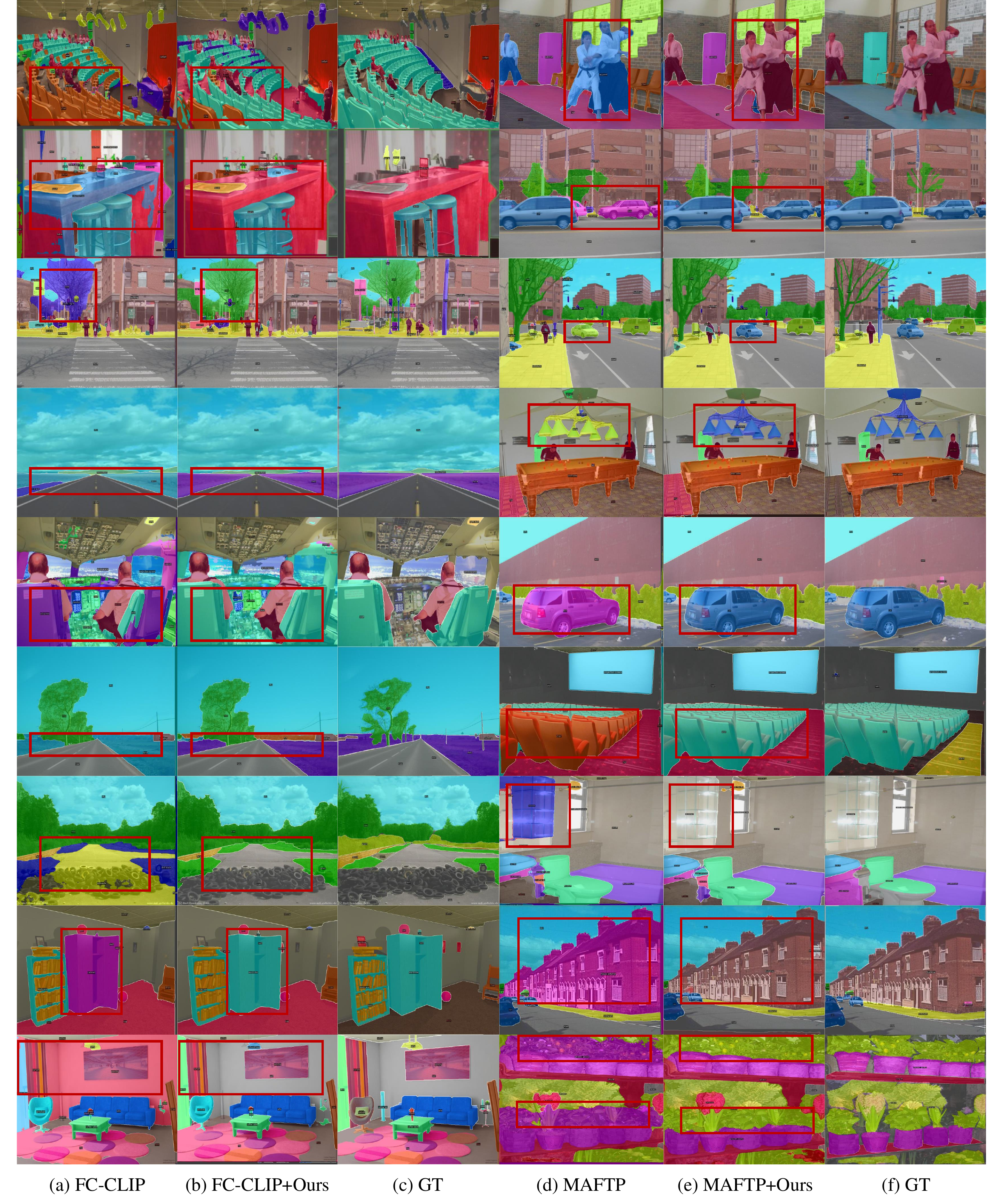}
    \caption{\textbf{Comparison of qualitative results on A-150~\cite{zhou2019semantic}.} We compare the segmentation results of existing open-vocabulary methods~\cite{yu2023convolutions,jiao2024collaborative} with those after integrating the Mask-Adapter.}
    \label{fig:vis_mask}
    \vspace{-5pt}
\end{figure*}

\begin{figure*}[htbp]
\centering
\includegraphics[width=1.00\linewidth,height=0.95\textheight]{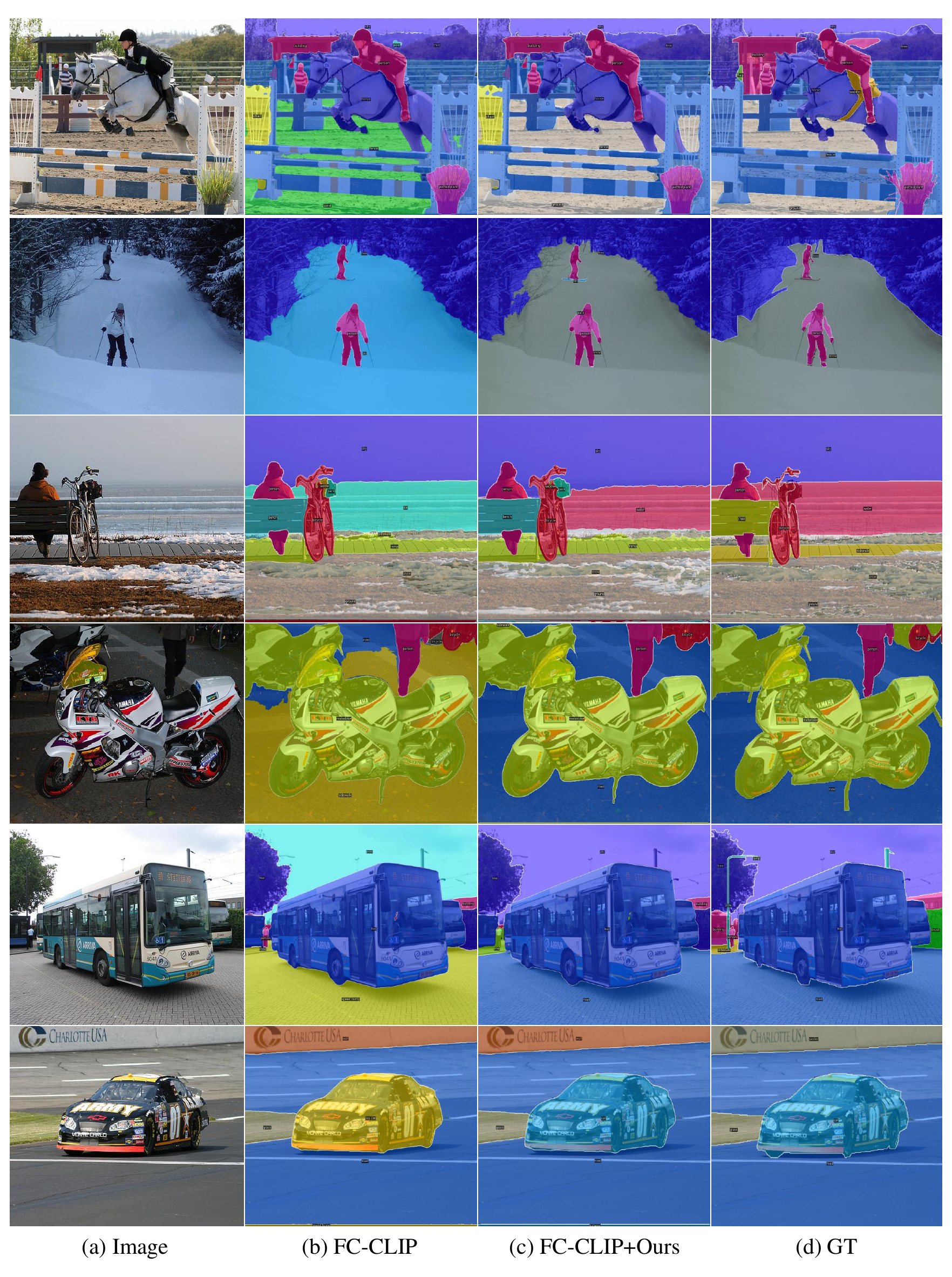}
    \caption{\textbf{Comparison of qualitative results on PC-459~\cite{mottaghi2014role}.} We compare the segmentation results of FC-CLIP~\cite{yu2023convolutions} and FC-CLIP integrated with Mask-Adapter on PC-459, highlighting the improvements achieved with Mask-Adapter.}
    \label{fig:vis_pc459}
\end{figure*}

\begin{figure*}[htbp]
    \centering
    \includegraphics[width=1.00\linewidth]{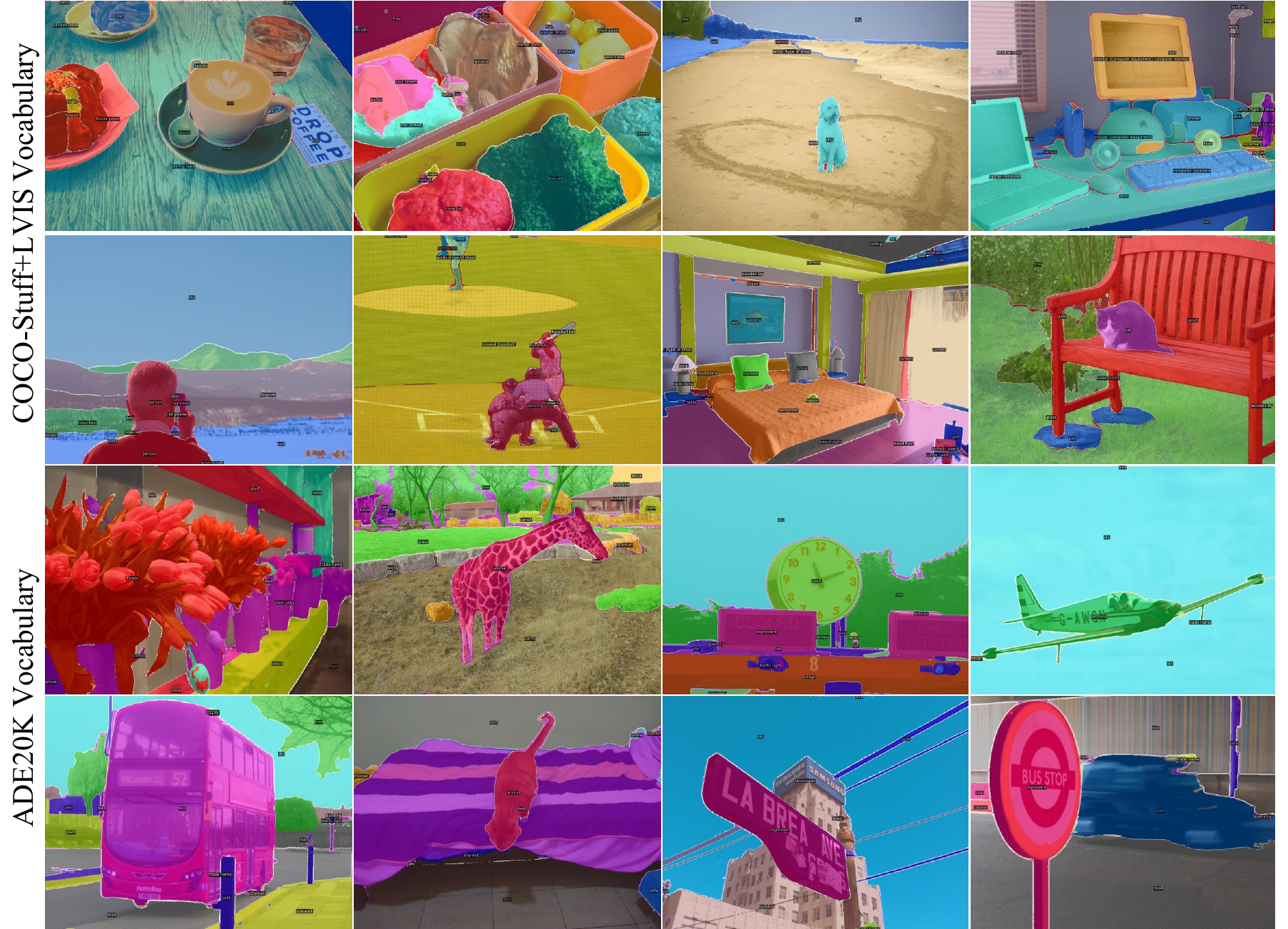}
    \caption{\textbf{Visualization results of Segment Anything with Mask-Adapter.} We present the segmentation results of Segment Anything integrated with Mask-Adapter on COCO-Stuff, using different vocabularies.}
    \label{fig:vis_sam}
    \vspace{-5pt}
\end{figure*}

We provide additional visualization results comparing segmentation performance on A-150~\cite{zhou2019semantic} using FC-CLIP~\cite{yu2023convolutions} and MAFTP~\cite{jiao2024collaborative}, as well as the results after integrating the Mask-Adapter (Fig.~\ref{fig:vis_mask}). Comparisons on PC-459~\cite{mottaghi2014role} are shown in Fig.~\ref{fig:vis_pc459}. Additionally, we demonstrate the performance of the combined SAM~\cite{kirillov2023segment} and Mask-Adapter on COCO-Stuff with different vocabularies in Fig.~\ref{fig:vis_sam}.

{
    \small
    \bibliographystyle{ieeenat_fullname}
    \bibliography{main}
}

\end{document}